\documentclass{article}
\usepackage{ijcai21}

\usepackage{times}
\usepackage{soul}
\usepackage{url}
\usepackage[hidelinks]{hyperref}
\usepackage[utf8]{inputenc}
\usepackage[small]{caption}
\usepackage{graphicx}
\usepackage{float}
\usepackage{subfigure}
\usepackage{amsmath}
\usepackage{amsthm}
\usepackage{booktabs}
\usepackage{algorithm}
\usepackage{algorithmicx}
\usepackage[noend]{algpseudocode}
\title{Adaptive Warm-Start MCTS in AlphaZero-like Deep Reinforcement Learning}

\author{
Hui Wang
\and
Mike Preuss\and
Aske Plaat
\affiliations
Leiden Institute of Advanced Computer Science, Leiden University,\\ Leiden, the Netherlands
\emails
h.wang.13@liacs.leidenuniv.nl
}
\begin{document}

\maketitle

\begin{abstract}
AlphaZero has achieved impressive performance in deep reinforcement learning by utilizing an architecture that combines 
search and training of a neural network in self-play. 
Many researchers are looking for ways to reproduce and improve results for other games/tasks. However, the architecture is designed to learn from scratch, tabula rasa,  accepting a cold-start problem in self-play.
Recently, a warm-start enhancement method for Monte Carlo Tree Search was proposed to improve the self-play starting phase. It employs a fixed parameter $I^\prime$ to control the warm-start length. Improved performance was reported in small board games. 
In this paper we present results with an adaptive switch method. Experiments show that our approach works better than the fixed $I^\prime$, especially for "deep," tactical, games (Othello and Connect Four). 
We conjecture that the adaptive value for $I^\prime$ is also influenced by the size of the game, and that on average $I^\prime$ will increase with game size. We conclude that AlphaZero-like deep reinforcement learning benefits from adaptive rollout based warm-start, as Rapid Action Value Estimate did for rollout-based reinforcement learning 15 years ago.  
\end{abstract}

\section{Introduction}\label{sec:introduction}
The combination of online Monte Carlo Tree Search (MCTS) \cite{browne2012survey} in self-play and offline neural network training has been widely applied as a deep reinforcement learning technique, in particular for solving game-related problems by means of the AlphaGo series programs~\cite{silver2016mastering,silver2017mastering,silver2018general}. The approach of this paradigm is to use game playing records from self-play by MCTS as training examples to train the neural network, whereas this trained neural network is used to inform the MCTS value and policy. Note that in contrast to AlphaGo Zero or AlphaZero,\footnote{AlphaZero is a general framework based on AlphaGo Zero for different games} the original AlphaGo also uses large amounts of expert data to train the neural network and a fast rollout policy together with the policy provided by neural network to guide the MCTS search. 

However, although the transition from a combination of using expert data and self-play (AlphaGo) to only using self-play (AlphaGo Zero and AlphaZero) appears to have only positive results, it does raise some questions.

The first question is: `should all human expert data be abandoned?' In other games we have  seen that human knowledge is essential for mastering complex games, such as StarCraft~\cite{vinyals2019grandmaster}. Then when should expert data be taken into consideration while training neural networks?

The second question is: `should the fast rollout policy be abandoned?' Recently, \cite{wang2020warm} have proposed to use warm-start search enhancements~\textbf{at the start phase} in AlphaZero-like self-play, which improves performance in 3 small board games. Instead of only using the neural network for value and policy, in the first few iterations, classic rollout can be used (or RAVE, or a combination, or a combination with the neural network). This can improve  training especially at the start phase of self-play training.

In fact, the essence of the warm-start search enhancement is to re-generate expert knowledge in the start phase of self-play training, to reduce the cold-start problem of playing against untrained agents. The method uses rollout (which can be seen as experts) instead of a randomly initialized neural network, up until a number of $I^\prime$ iterations, when it switches to the regular value network. 
In their experiments, the $I^\prime$ was fixed at 5.
Obviously, a fixed $I^\prime$ may not be optimal. Therefore, in this work, we propose an adaptive switch method. The method uses an \textbf{arena} in the self-play stage~(see Algorithm~\ref{alg:adaptivea0g}), where the search enhancement and the default MCTS are matched, to judge whether to switch or not. With this mechanism, we can dynamically switch off the enhancement if it is no longer better than the default MCTS player, as the neural network is being trained. 

Our main contributions can be summarized as follows:

\begin{enumerate}
\item Warm-start method improves the Elo of AlphaZero-like self-play in small games, but it
 introduces a new hyper-parameter. Adaptive warm-start further improves performance and removes the hyper-parameter.
\item For deep games (with a small branching factor) warm-start works better than for shallow games. This indicates that the effectiveness of warm-start method may increase for larger games.




\end{enumerate}

The rest of paper is designed as follows. 
An overview of the most relevant literature is given in Sect.\,\ref{sec:relatedwork}. Before proposing our adaptive switch method in Sect.\,\ref{sec:adaptivealphazero}, we describe the warm-start AlphaZero-like self-play algorithm in Sect.\,\ref{sec:warmalphazero}. Thereafter, we set up the experiments in Sect.\,\ref{sec:setup} and present their results in Sect.\,\ref{sec:results}. Finally, we conclude our paper and discuss future work.

\section{Related Work}\label{sec:relatedwork}

There are a lot of early successful works in reinforcement learning~\cite{sutton2018reinforcement}, e.g.  using temporal difference learning with a neural network to play backgammon \cite{tesauro1995temporal}. MCTS has also been well studied, and many variants/enhancements were designed to solve problems in the domain of sequential decisions, especially on games. For example, enhancements such as Rapid Action Value Estimate~(RAVE) and All Moves as First~(AMAF) have been conceived to improve MCTS~\cite{gelly2007combining,gelly2011monte}. 
The AlphaGo series algorithms replace the table based model with a deep neural network based model, where the neural network has a policy head~(for evaluating of a state) and a value head~(for learning a best action)~\cite{wang2019alternative}, enabled by the GPU hardware development. Thereafter, the structure that combines MCTS with neural network training has become a typical approach for reinforcement learning tasks and many successful applications~\cite{segler2018planning,wang2020tackling} of this kind model-based deep reinforcement learning~\cite{schmidhuber2015deep}. Comparing AlphaGo with AlphaGo Zero and AlphaZero, the latter did not use any expert data to train the neural network, and abandoned the fast rollout policy for improving the MCTS on the trained neural network. Therefore, all training data is generated purely by self-play, which is also a very important feature of reinforcement learning. We base our work on an open reimplementation of AlphaZero, AlphaZero General~\cite{surag2018}.

There are many interesting works on self-play in reinforcement learning ~\cite{tesauro1995temporal,runarsson2005coevolution,plaat2020learning}. Temporal difference learning for acquiring position evaluation in small board Go with co-evolution has been compared to self-play~\cite{runarsson2005coevolution}. These works demonstrated the impressive results for self-play and emphasized its importance.

Within a general game playing framework, in order to improve training examples efficiency,~\cite{wang2018assessing} assessed the potential of classical Q-learning by introducing Monte Carlo Search enhancements. In an AlphaZero-like self-play framework,~\cite{wu2019accelerating} used domain-specific features and optimizations, starting from random initialization and no preexisting data, to accelerate the training. 

However, AlphaStar, the acclaimed algorithm for beating human professionals at StarCraft \cite{vinyals2019grandmaster}, went back to utilizing human expert data, thereby suggesting that this is still an option at the start phase of training. Apart from this, there are few studies on applying MCTS enhancements in AlphaZero-like self-play.  
Only \cite{wang2020warm}, which proposed a warm-start search enhancement method, pointed out the promising potential of utilizing MCTS enhancements~(like a rollout policy) to re-generate expert data at the start phase of training. Our approach  differs from AlphaStar, as we generate expert data using MCTS enhancements other than collecting it from humans; further, compared to the static warm-start of \cite{wang2020warm}, we propose an adaptive method to control the iteration length of using such enhancements instead of a fixed $I^\prime$.

\section{Warm-Start AlphaZero Self-play}\label{sec:warmalphazero}
We will now introduce the warm-start enhancement method. 
\subsection{The Algorithm Framework}\label{a0gintroduction}

Based on~\cite{silver2018general,wang2019alternative,wang2020warm}, the core of AlphaZero-like self-play~(see Algorithm~\ref{alg:a0gwarm}) is an iterative loop which consists of three different stages within the single iteration as follows:
\begin{enumerate}
    \item \textbf{self-play}: The first stage is  playing several games against with itself to generate training examples. 
    \item \textbf{neural network training}: The second stage is feeding the neural network with training examples~(generated in the first stage) to train a new model. 
    \item \textbf{arena comparison}: The last stage is employing a tournament to compare the newly trained model and the old model to decide whether to update or not.   
\end{enumerate}

The detail description of these 3 stages can be found in~\cite{wang2020warm}. Note that in the Algorithm~\ref{alg:a0gwarm}, line~\ref{linewithorwithoutenhancement}, a fixed $I^\prime$ is employed to control whether to use neural network  MCTS or MCTS enhancements, the $I^\prime$ should be set as relatively smaller than $I$, which is known as warm-start search. The MCTS algorithm and MCTS enhancements will be introduced in next subsections.

\begin{algorithm}[bth!]
\caption{Warm-start AlphaZero-like Self-play Algorithm}
\label{alg:a0gwarm}
\begin{algorithmic}[1]
\footnotesize
\State Randomly initialize $f_\theta$, assign retrain buffer $D$
\For{iteration=1, $\dots$,$I^\prime$, $\dots$, $I$}
\For{episode=1,$\dots$, $E$}\Comment{self-play}
\For{t=1, $\dots$, $T^\prime$, $\dots$, $T$} 
\If{$I \leq I'$} $\pi_{t} \leftarrow$ \textbf{MCTS Enhancement}\label{linewithorwithoutenhancement}
\Else{} $\pi_{t} \leftarrow$ \textbf{default MCTS} 
\EndIf
\If{$t \leq T'$} $a_t=$ randomly select on $\pi_t$ 
\Else{} $a_t = \arg\max_a(\pi_t)$ 
\EndIf
\State executeAction($s_t$, $a_t$)
\EndFor
\State $D \leftarrow (s_t,\pi_t,z_t)$ with  outcome $z_{t\in [1,T]}$
\EndFor
\State Sample  minibatch~($s_j$, $\pi_j$, $z_j$) from $D$ \Comment{training}
\State Train $f_{\theta^\prime}\leftarrow f_\theta$ 

\State $f_\theta=f_{\theta^\prime}$ if $f_{\theta^\prime}$ is better, using \textbf{default MCTS}\Comment{arena}
\EndFor
\State \Return $f_\theta$;
\end{algorithmic}
\end{algorithm}

\subsection{MCTS}\label{subsec:mctsnnt}
Classical MCTS has shown successful performance to solve complex games, by taking random samples in the search space to evaluate the state value. Basically, the classical MCTS algorithm can be divided into 4 stages, which are known as \textit{selection}, \textit{expansion}, \textit{rollout} and \textit{backpropagate}~\cite{browne2012survey}. However, for the default MCTS in AlphaZero-like self-play~(eg. our Baseline), the neural network directly informs the MCTS state policy and value to guide the search instead of running a rollout.

\subsection{MCTS enhancements}\label{subsec:mctsenhancements}
In this paper, we adopt the same two individual enhancements and three combinations to improve neural network training as were used by~\cite{wang2020warm}.

\noindent \textbf{Rollout} is running a classic MCTS random rollout to get a value that provides more meaningful information than a value from random initialized neural network.

\noindent \textbf{RAVE} is a well-studied enhancement to cope with the cold-start of MCTS in games like Go~\cite{gelly2007combining}, where the playout-sequence can be transposed. The core idea of RAVE is using AMAF to update the state visit count $N_{rave}$ and Q-value $Q_{rave}$, which are written as: $N_{rave}(s_{t_1},a_{t_2})\leftarrow N_{rave}(s_{t_1},a_{t_2})+1$, $Q_{rave}(s_{t_1},a_{t_2})\leftarrow\frac{N_{rave}(s_{t_1},a_{t_2})*Q_{rave}(s_{t_1},a_{t_2})+v}{N_{rave}(s_{t_1},a_{t_2})+1}$, where $s_{t_1}\in VisitedPath$, and $a_{t_2}\in A(s_{t_1})$, and for $\forall t < t_2, a_t\neq a_{t_2}$. The P-UCT of RAVE is calculated as follows: 

\begin{equation}\label{equationUCTrave}
 UCT_{rave}(s,a)=(1-\beta)* U(s,a) + \beta*U_{rave}(s,a)
 \end{equation}
 where 
 \begin{equation}
 U_{rave}(s,a) = Q_{rave}(s,a) + c*P(s,a)\frac{\sqrt{N_{rave}(s,\cdot)}}{N_{rave}(s,a)+1}
 \end{equation}
 and 
 \begin{equation}
 \beta=\sqrt{\frac{equivalence}{3*N(s,\cdot)+equivalence}}
 \end{equation}
 The value of equivalence is usually set to the number of MCTS simulations~(i.e $m$=100 in our experiments).

\noindent \textbf{RoRa} is the combination which simply adds the random rollout to enhance RAVE.

\noindent \textbf{WRo} introduces a weighted sum of rollout value and the neural network value as the return value to guide MCTS. In our experiments, $v(s)$ is computed as follows:
\begin{equation}
v(s)=(1-weight)*v_{network}+ weight*v_{rollout}
\end{equation}

\noindent \textbf{WRoRa} also employs a weighted sum to combine the value from the neural network and the value of RoRa. The $v(s)$ for MCTS search in WRoRa is computed as follows:
\begin{equation}
v(s)=(1-weight)*v_{network}+ weight*v_{rora}
\end{equation}

Different from~\cite{wang2020warm}, since there is no pre-determined  $I^\prime$, in our work, $weight$ is simply calculated as $1/i, i\in [1,I]$, where $i$ is the current iteration number.

\section{Adaptive Warm-Start Switch Method}\label{sec:adaptivealphazero}

The fixed $I^\prime$  to control the length of using warm-start search enhancements as suggested by \cite{wang2020warm} works, but seems to require different parameter values for different games. In consequence, a costly tuning process would be necessary for each game. Thus, an adaptive method would have multiple advantages.

We notice that the core of the warm-start method is re-generating expert data to train the neural network at the start phase of self-training to avoid learning from weak (random or near random) self-play. We suggest to stop the warm-start when the neural network is on average playing stronger  than the enhancements. Therefore, in the self-play, we employ a tournament to compare the standard AlphaZero-like self-play model~(Baseline) and the enhancements~(see Algorithm~\ref{alg:adaptivea0g}). The switch occurs once the Baseline MCTS wins more than 50\%. In order to avoid spending too much time on this, these arena game records will directly be used as training examples, indicating that the training data is played by the enhancements and the Baseline. This scheme enables to switch at individual points in time for different games and even different training runs.

\begin{algorithm}[bth!]
\caption{Adaptive Warm-Start Switch Algorithm}
\label{alg:adaptivea0g}
\begin{algorithmic}[1]
\footnotesize
\State Initialize $f_\theta$ with random weights; Initialize retrain buffer $D$, Switch$\leftarrow$False, $r_{mcts} \leftarrow 0$
\For{iteration=1, $\dots$, $I$}
\Comment no $I^\prime$
\If{not Switch}\Comment{not switch}
\For{episode=1,$\dots$, $E$}\Comment{arena with enhancements}
\For{t=1, $\dots$, $T^\prime$, $\dots$, $T$}
\If{$episode \leq E/2$} 
\If{t is odd} $\pi_{t} \leftarrow$
\textbf{MCTS Enhancement}
\Else{}  $\pi_{t} \leftarrow$  \textbf{default MCTS} 
\EndIf
\Else{} 
\If{t is odd}
$\pi_{t} \leftarrow$ \textbf{default MCTS} 
\Else{} $\pi_{t} \leftarrow$ \textbf{MCTS Enhancement}
\EndIf
\EndIf
\If{$t \leq T'$} $a_t=$ randomly select on $\pi_t$ 
\Else{} $a_t = \arg\max_a(\pi_t)$ 
\EndIf

\State executeAction($s_t$, $a_t$)
\EndFor
\State $D \leftarrow (s_t,\pi_t,z_t)$ with  outcome $z_{t\in [1,T]}$
\State $r_{mcts}$+= reward of \textbf{default MCTS} in this episode
\EndFor
\Else\Comment{switch}
\For{episode=1,$\dots$, $E$}\Comment{purely self-play}
\For{t=1, $\dots$, $T^\prime$, $\dots$, $T$} 
\State$\pi_{t} \leftarrow$ 
\textbf{default MCTS}
\If{$t \leq T'$} $a_t=$ randomly select on $\pi_t$ 
\Else{} $a_t = \arg\max_a(\pi_t)$ 
\EndIf
\State executeAction($s_t$, $a_t$)

\EndFor
\State $D \leftarrow (s_t,\pi_t,z_t)$ with  outcome $z_{t\in [1,T]}$
\EndFor
\EndIf
\State Set Switch$\leftarrow$True if $r_{mcts}>$0, and set $r_{mcts} \leftarrow 0$
\State Sample  minibatch~($s_j$, $\pi_j$, $z_j$) from $D$ \Comment{training}
\State Train $f_{\theta^\prime}\leftarrow f_\theta$ 

\State $f_\theta=f_{\theta^\prime}$ if $f_{\theta^\prime}$ is better, using \textbf{default MCTS} \Comment{arena}
\EndFor
\State \Return $f_\theta$;
\end{algorithmic}
\end{algorithm}

\section{Experimental Setup}\label{sec:setup}

Since~\cite{wang2020warm} only studied the winrate of single rollout and RAVE against a  random player, this can be used as a test to check whether rollout and RAVE work. However, it does not reveal any information about relative playing strength, which is necessary to explain how good training examples provided by MCTS enhancements actually are. Therefore, at first we let all 5 enhancements and the baseline MCTS play 100 repetitions with each other on the same 3 games~(6$\times$6 Connect Four, Othello and Gobang, game description can be found in~\cite{wang2020warm}) in order to investigate the relative playing strength of each pair.

In the second experiment, we tune the fixed $I^\prime$, where $I^\prime\in \{1,3,5,7,9\}$, for different search enhancements, based on Algorithm~\ref{alg:a0gwarm} to play 6$\times$6 Connect Four.

In our last experiment, we use new adaptive switch method~Algorithm~\ref{alg:adaptivea0g} to play 6$\times$6 Othello, Connect Four and Gobang. We set parameters values according to Table~\ref{defaulttab}. The parameter choices are based on~\cite{wang2020analysis}.

Our experiments are run on a high-performance computing (HPC) server, which is a  cluster consisting of 20 CPU nodes (40 TFlops) and 10 GPU nodes (40 GPU, 20 TFlops CPU + 536 TFlops GPU).  We use small versions of games (6$\times$6) in order to perform a medium number of repetitions. In the following, our figures show error bars of 8 runs, of 100 iterations of self-play. Each single run is deployed in a single GPU which takes several days for different games.

\begin{table}[bht!]
\centering\hspace*{-2.3em}
\begin{tabular}{lll}
\hline
Parameter& Description & Value\\
\hline
\emph{I}	&number of iteration		&100 \\
\emph{rs}	& number of retrain iteration		&20\\

\emph{ep}	& number of epoch	&10	\\

\emph{E}    &number of episode		&50\\
\emph{bs}	& batch size	&64\\

\emph{T'}	&step threshold		&15\\
\emph{lr}	& learning rate	&	0.005\\

\emph{m}	&MCTS simulation times	&100\\
\emph{d}    & dropout probability &0.3\\

\emph{c}    &weight in UCT		&1.0\\
\emph{n}   & number of comparison games	&40\\

\emph{u}	& update threshold		&0.6\\
\hline
\end{tabular}
\caption{Default Parameter Setting}\label{defaulttab}
\end{table}

\section{Results}\label{sec:results}
We list results for a tournament of Baseline and enhancements~(Table~\ref{comparisontable}). Digging deeper, we also report the effect of the hyper-parameter $I'$ (Fig~\ref{fig:subfigdifferentheuristics}). And results for the adaptive warm-start switch   are shown in Table~\ref{tab:Ifordifferentgames}, Fig~\ref{fig:balanceofothello} and Fig~\ref{fig:fixedvsadapt}.

\subsection{MCTS vs MCTS Enhancements}

Here, we compare the Baseline player (the neural network is initialized randomly which can be regarded as an arena in the first iteration self-play) to the other 5 MCTS enhancements players on 3 different games.
Each pair performs 100 repetitions.  In Table~\ref{comparisontable}, we can see that for Connect Four, the highest winrate is achieved by WRoRa, the lowest by Rave. Except Rave, others are all higher than 50\%, showing that the enhancements~(except Rave) are better than the untrained Baseline. In Gobang, it is similar, Rave is the lowest, RoRa is the highest. But the winrates are relatively lower than that in other 2 games. It is interesting that in Othello, all winrates are relatively the highest compared to the 2 other games~(nearly 100\%), although Rave still achieves the lowest winrate which is higher than 50\%. 

One reason that enhancements work best in Othello is that the Othello game tree is the longest and narrowest (low branching factor). Enhancements like Rollout can provide relatively accurate estimations for these trees. In contrast, Gobang has the shortest game length and the most legal action options. Enhancements like Rollout do not contribute much to the search in short but wide search tree with limited MCTS simulation. As in shorter games it is more likely to reach a terminal state, both Baseline and enhancements get the true result. Therefore, in comparison to MCTS, enhancements like Rollout work better while it does not terminate too fast. Rave is filling more state action pairs based on information from the neural network, its weaknesses at the beginning are more emphasized. After some iterations of training, the neural network becomes smarter, and Rave can therefore enhance the performance as shown in~\cite{wang2020warm}.

\begin{table}[tbh!]
\centering
\caption{Results of comparing default MCTS with Rollout, Rave, RoRa, WRo and WRoRa, respectively on the three games with random neural network, weight as 1/2, $T^\prime$=0, win rates in percent (row vs column), 100 repetitions each. }\label{comparisontable}
\begin{tabular}{llll}
\hline
&\multicolumn{3}{c}{Default MCTS} \\
\hline
&ConnectFour&Othello&Gobang\\
\hline
Rollout 	&64  &  93 &65	 \\

Rave &27.5	 &53& 43\\

RoRa &76	  &98  &70\\
WRo &82	  &96& 57\\
WRoRa&	82.5  & 99& 62\\
\hline
\end{tabular}
\end{table}

\subsection{Fixed $I^\prime$ Tuning}
\begin{figure}[!tbh]
\centering
\hspace*{-1.5em}
\subfigure[Rollout]{\label{fig:subfig:Rollout}
\includegraphics[width=0.53\columnwidth]{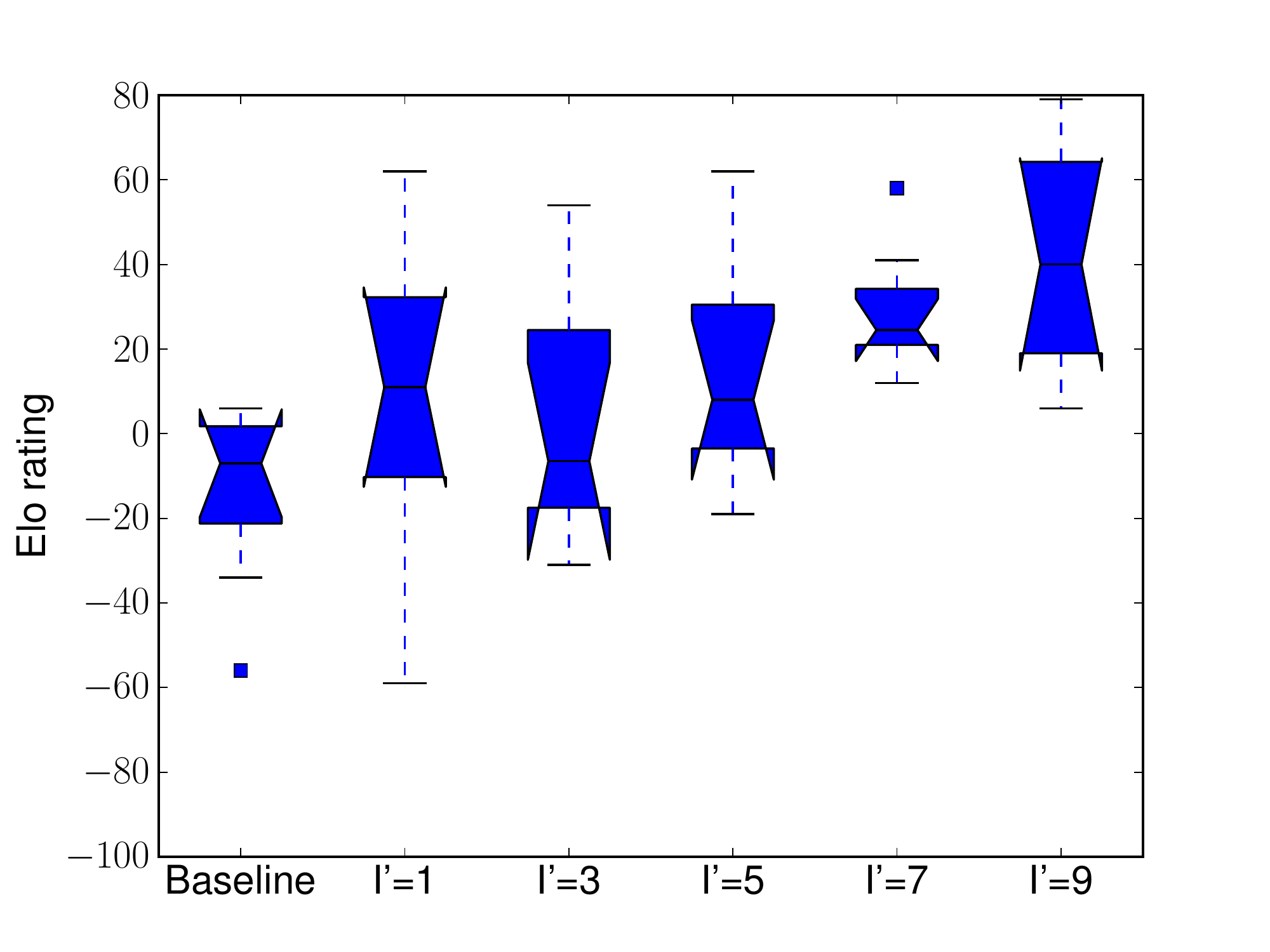}}
\hspace*{-1.5em}
\subfigure[Rave]{\label{fig:subfig:Rave}
\includegraphics[width=0.53\columnwidth]{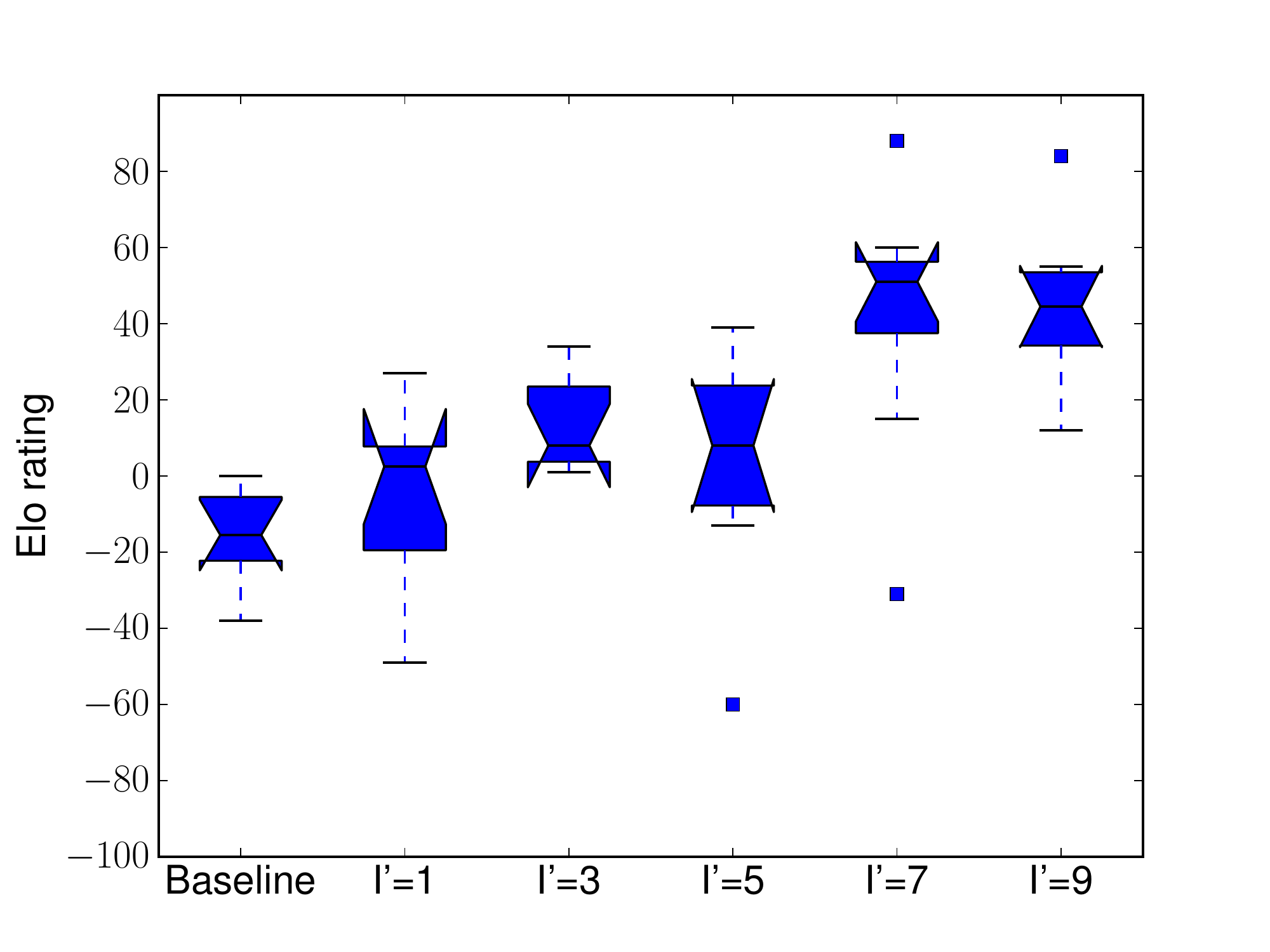}}
\hspace*{-1.5em}
\subfigure[RoRa]{\label{fig:subfig:RoRa}
\includegraphics[width=0.53\columnwidth]{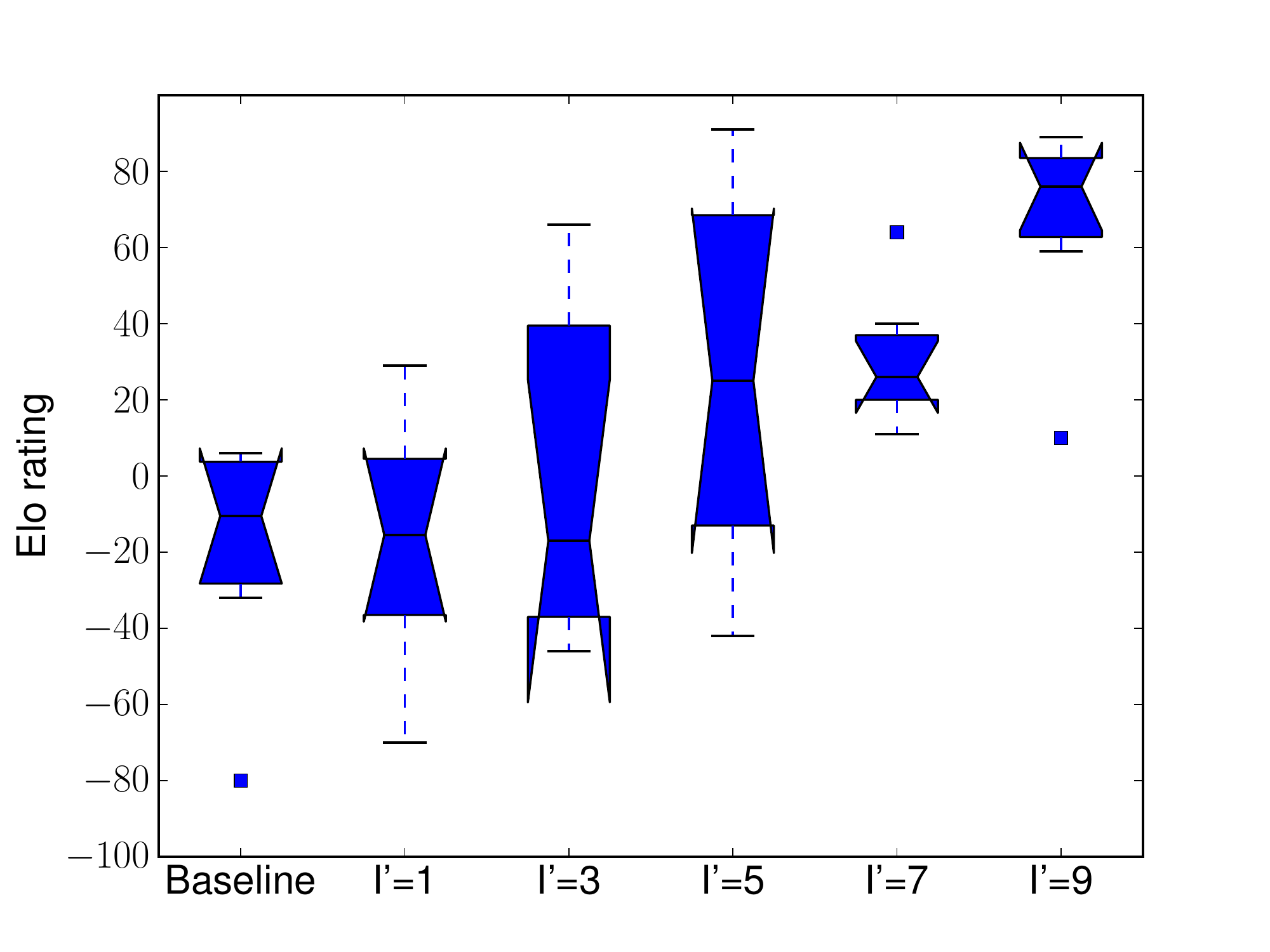}}
\hspace*{-1.5em}
\subfigure[WRo]{\label{fig:subfig:WRo}
\includegraphics[width=0.53\columnwidth]{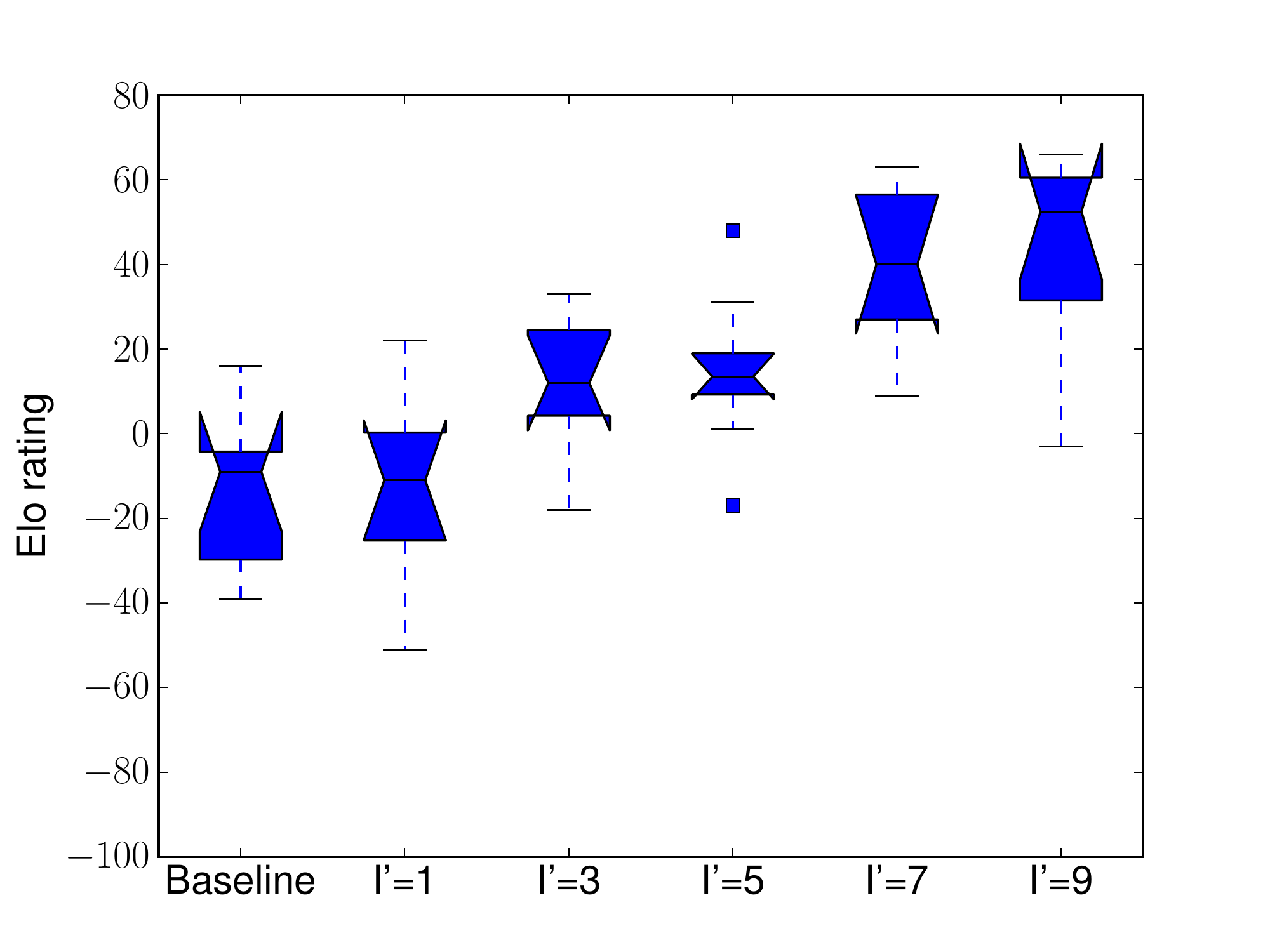}}
\hspace*{-1.5em}
\subfigure[WRoRa]{\label{fig:subfig:WRoRa}
\includegraphics[width=0.53\columnwidth]{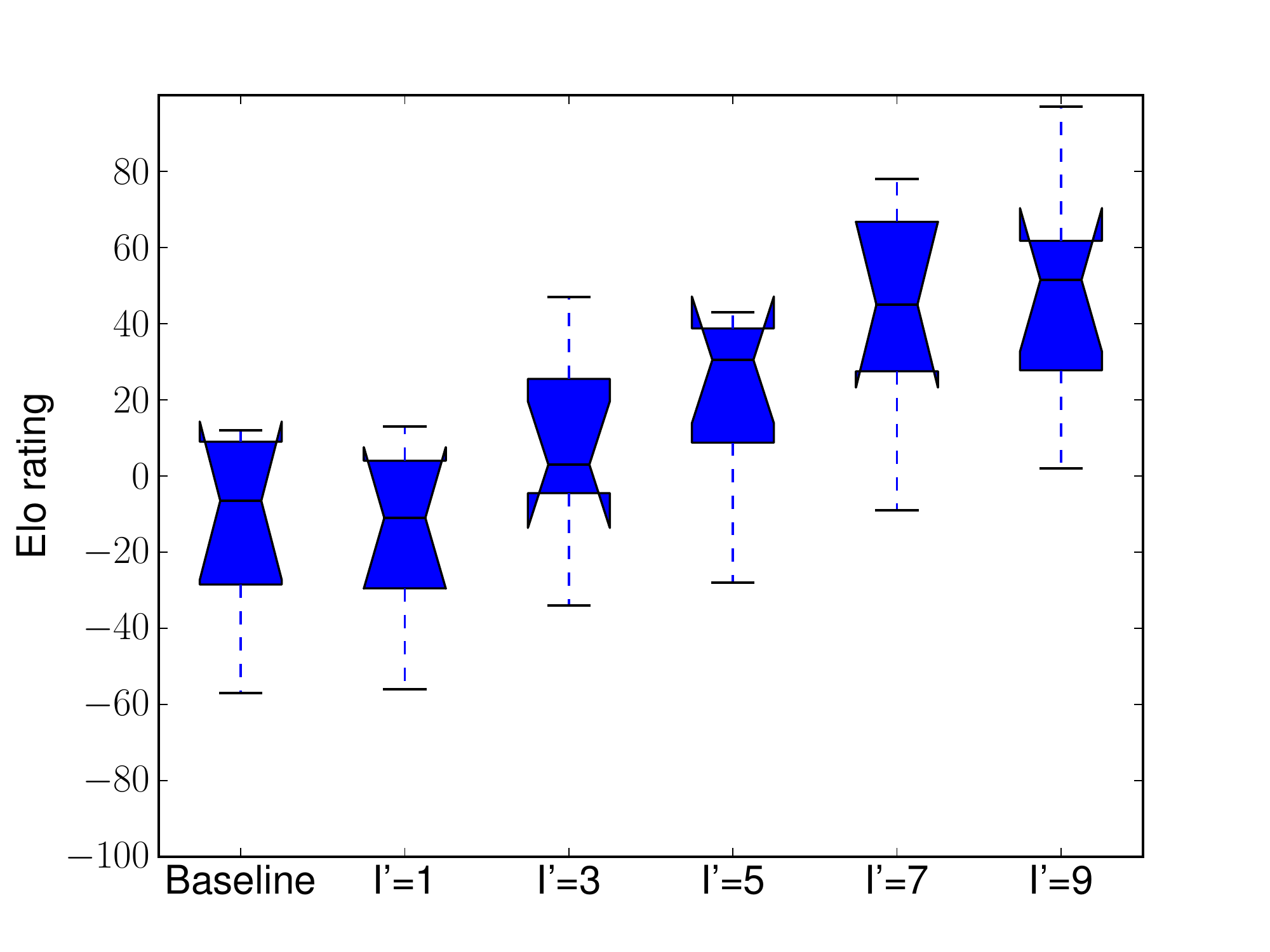}}
\caption{Elo ratings for different warm-start phase iterations with different search enhancement on 6$\times$6 Connect Four}
\label{fig:subfigdifferentheuristics} 
\end{figure}

Taking Connect Four as an example, in this experiment we search for an optimal fixed $I^\prime$ value, utilizing the warm-start search method proposed in~\cite{wang2020warm}. We set $I^\prime$ as 1, 3, 5, 7, 9 respectively~(the value should be relatively small since the enhancement is only expected to be used at the start phase of training). The Elo ratings of each enhancements using different $I^\prime$ are presented in Fig~\ref{fig:subfigdifferentheuristics}. The Elo ratings are calculated based on the tournament results using a Bayesian Elo computation system~\cite{coulom2008whole}, same for Fig~\ref{fig:fixedvsadapt}.  We can see that for Rave and WRoRa, it turns out that $I^\prime=7$ is the optimal value for fixed $I^\prime$ warm-start framework, for others, it is still unclear which value is the best, indicating that the tuning is inefficient and costly.

\subsection{Adaptive Warm-Start Switch}

In this final experiment, we apply the newly suggested adaptive switch warm-start search enhancement method and compare it to the fixed $I^\prime$. We are especially interested in the averages and variances of the switching times that result from adaptive switching.

 We train models with the parameters in Table~\ref{defaulttab} and then let them compete against each other in different games. In addition, we record the specific iteration number where the switch occurs for every training run and the corresponding self-play arena rewards of MCTS before this iteration. A statistic of the iteration number for 3 games is  shown in Table~\ref{tab:Ifordifferentgames}.

\begin{table}[!bth]
\centering
\caption{Switching iterations for training on different games with different enhancements over 8 repetitions~(average iteration number $\pm$ standard deviation)}
\begin{tabular}{llll}
\hline
& Connect Four&Othello &Gobang \\
\hline
Rollout&6.625$\pm$ 3.039&5.5$\pm$ 1.732&1.375$\pm$0.484\\
Rave&2.375$\pm$1.218&3.125 $\pm$2.667&1.125$\pm$0.331\\
RoRa&7.75 $\pm$4.74&5.125 $\pm$1.364&1.125$\pm$0.331\\
WRo&4.25$\pm$1.561&4.375$\pm$1.654&1.125$\pm$0.331\\
WRoRa&4.375$\pm$1.576&4.0$\pm$1.0&1.25$\pm$0.433\\
\hline
\end{tabular}
\label{tab:Ifordifferentgames}
\end{table}

The table shows that, generally, the iteration number is relatively small compared to the total length of the training (100 iterations), and in these small games the neural network is quickly getting stronger. Besides, not only for different games, the switch iteration is different, but also for different training runs on the same game, the switch iteration also varies. This is because for different training runs, the neural network training progresses differently (we already start from different random initializations). Therefore, a fixed $I^\prime$ can not be used for each specific training. Note that for Gobang, a game with a large branching factor, with the default setting, it always switches at the first iteration. Therefore, we also test with larger $m = 200$, thereby providing more time to the MCTS. With this change, there are several runs keeping the enhancements~see Table~\ref{tab:Ifordifferentgames}, but it still shows a small influence on this game.

\begin{figure}[!b]
\centering
\hspace*{-1.5em}
\subfigure[Rollout]{\label{fig:subfig:balancerolloutothello}
\includegraphics[width=0.53\columnwidth]{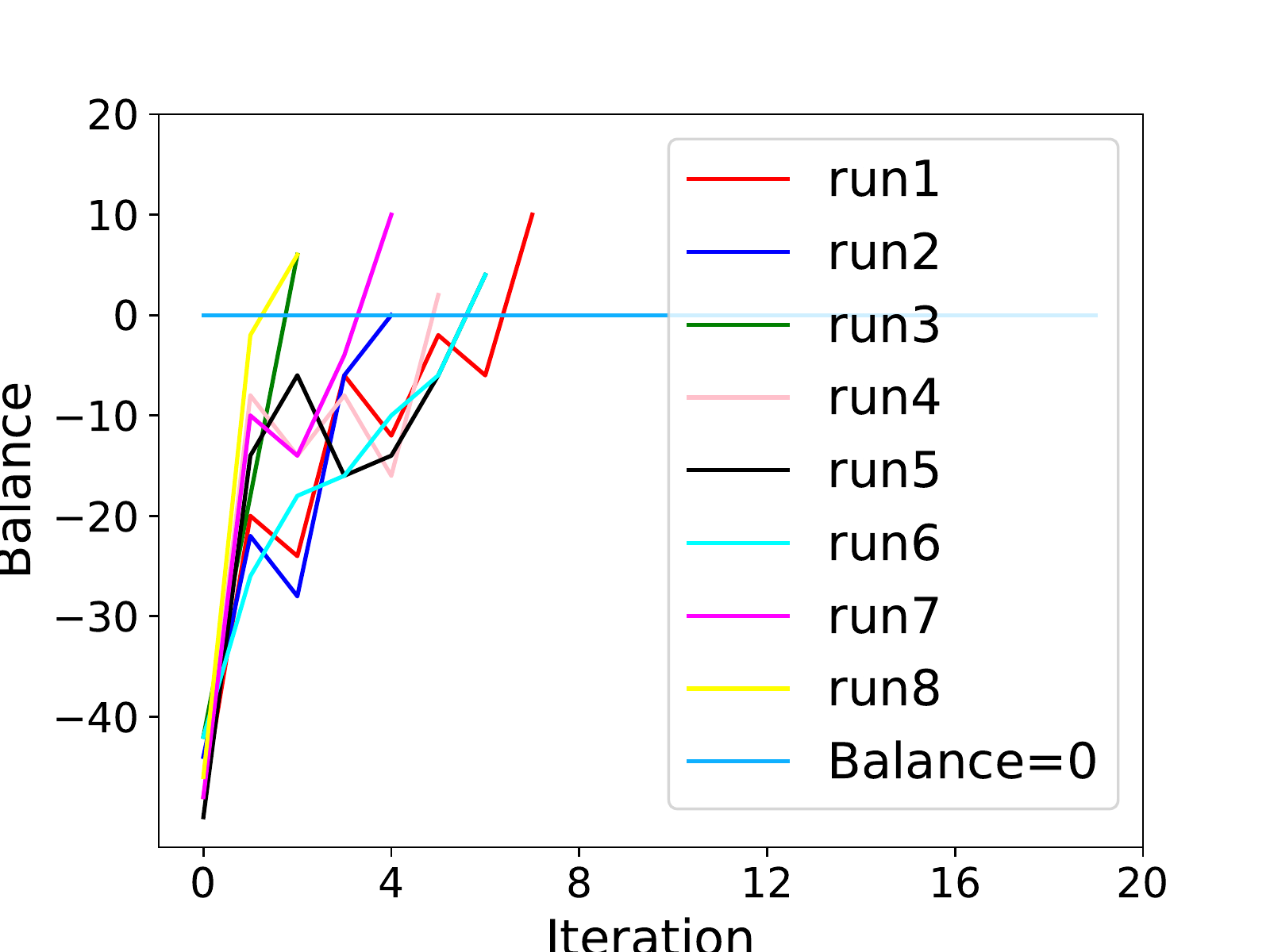}}
\hspace*{-1.5em}
\subfigure[Rave]{\label{fig:subfig:balanceraveothello}
\includegraphics[width=0.53\columnwidth]{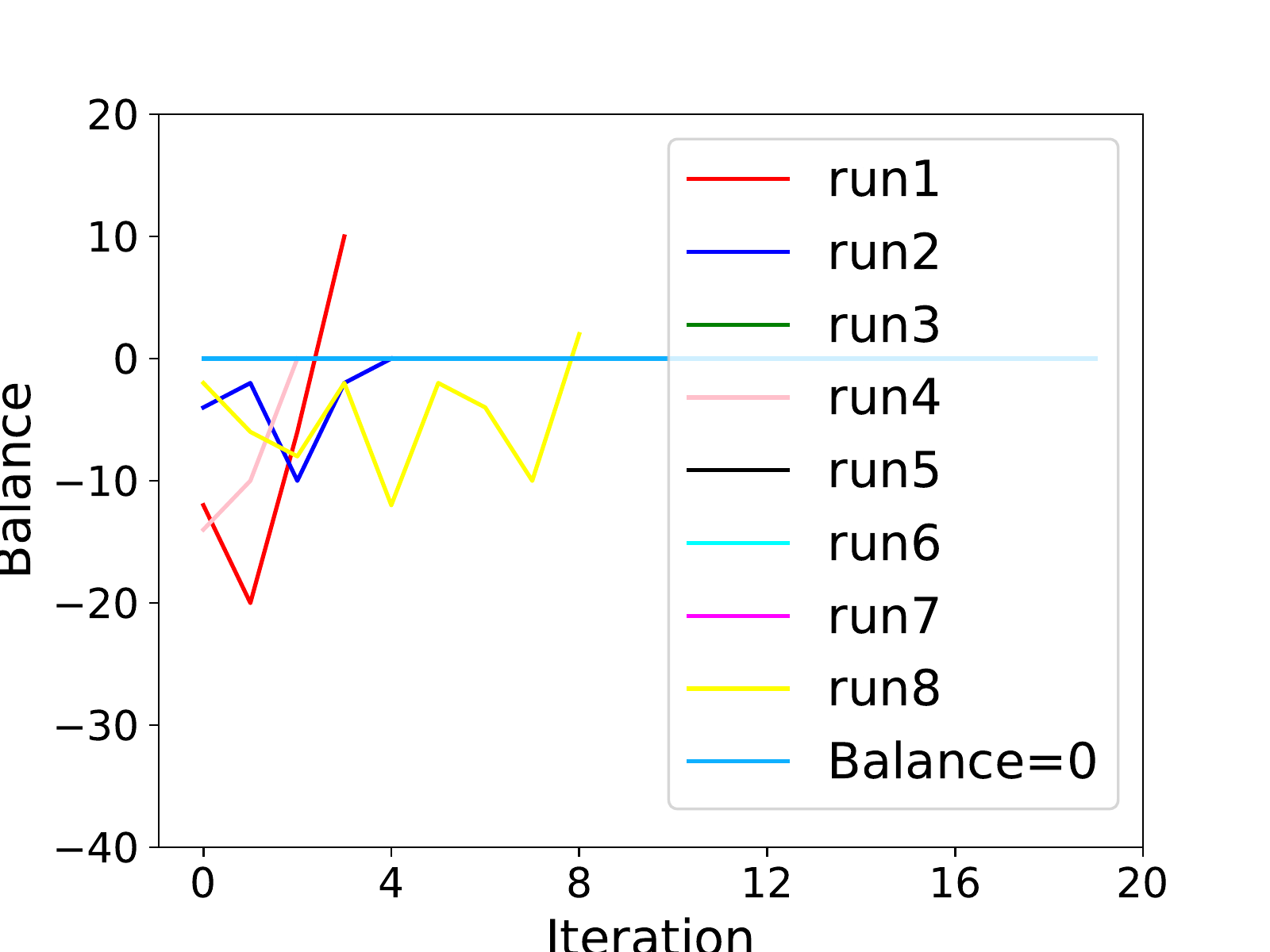}}
\hspace*{-1.5em}
\subfigure[RoRa]{\label{fig:subfig:balanceroraothello}
\includegraphics[width=0.53\columnwidth]{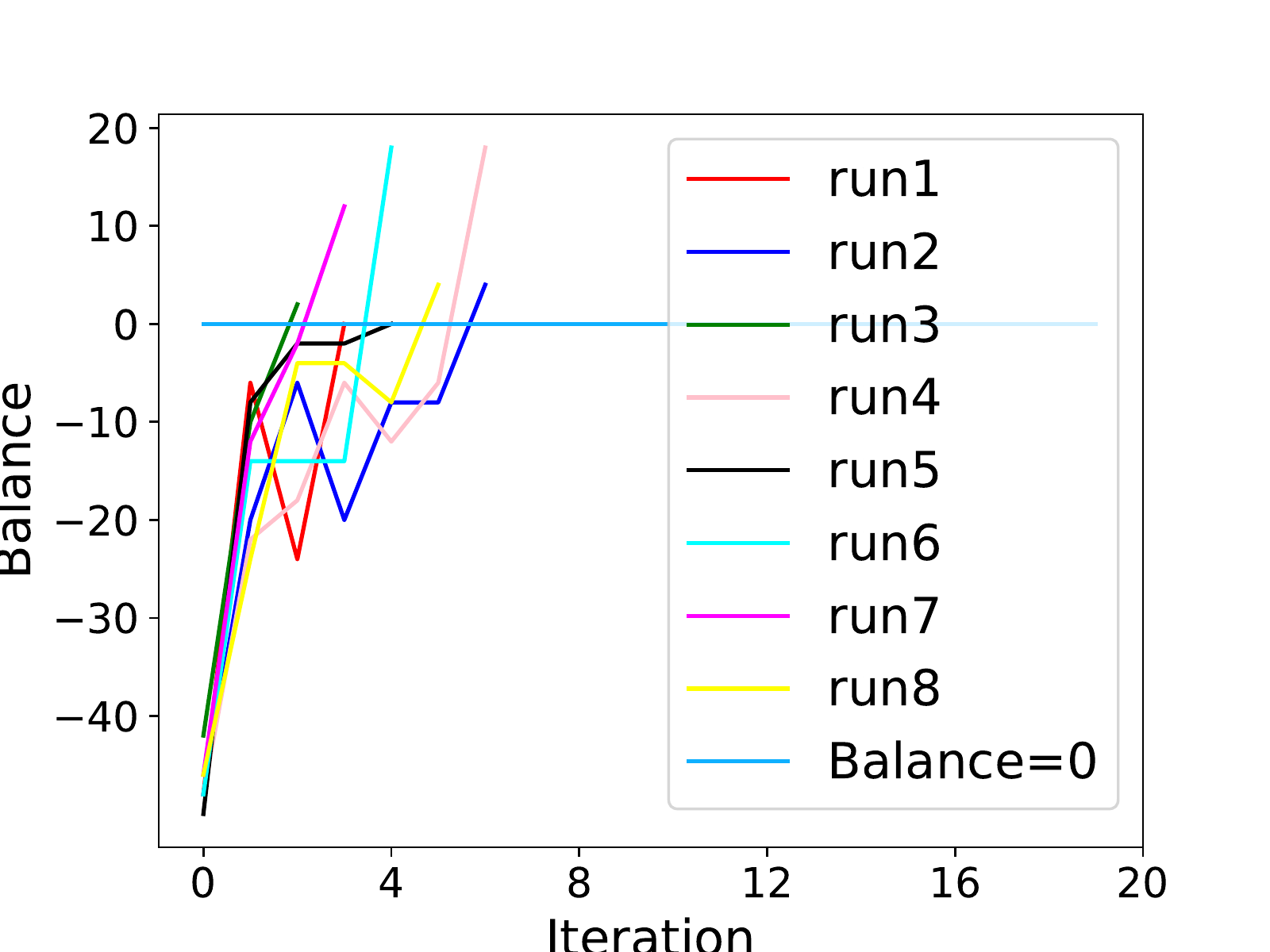}}
\hspace*{-1.5em}
\subfigure[WRo]{\label{fig:subfig:balancewroothello}
\includegraphics[width=0.53\columnwidth]{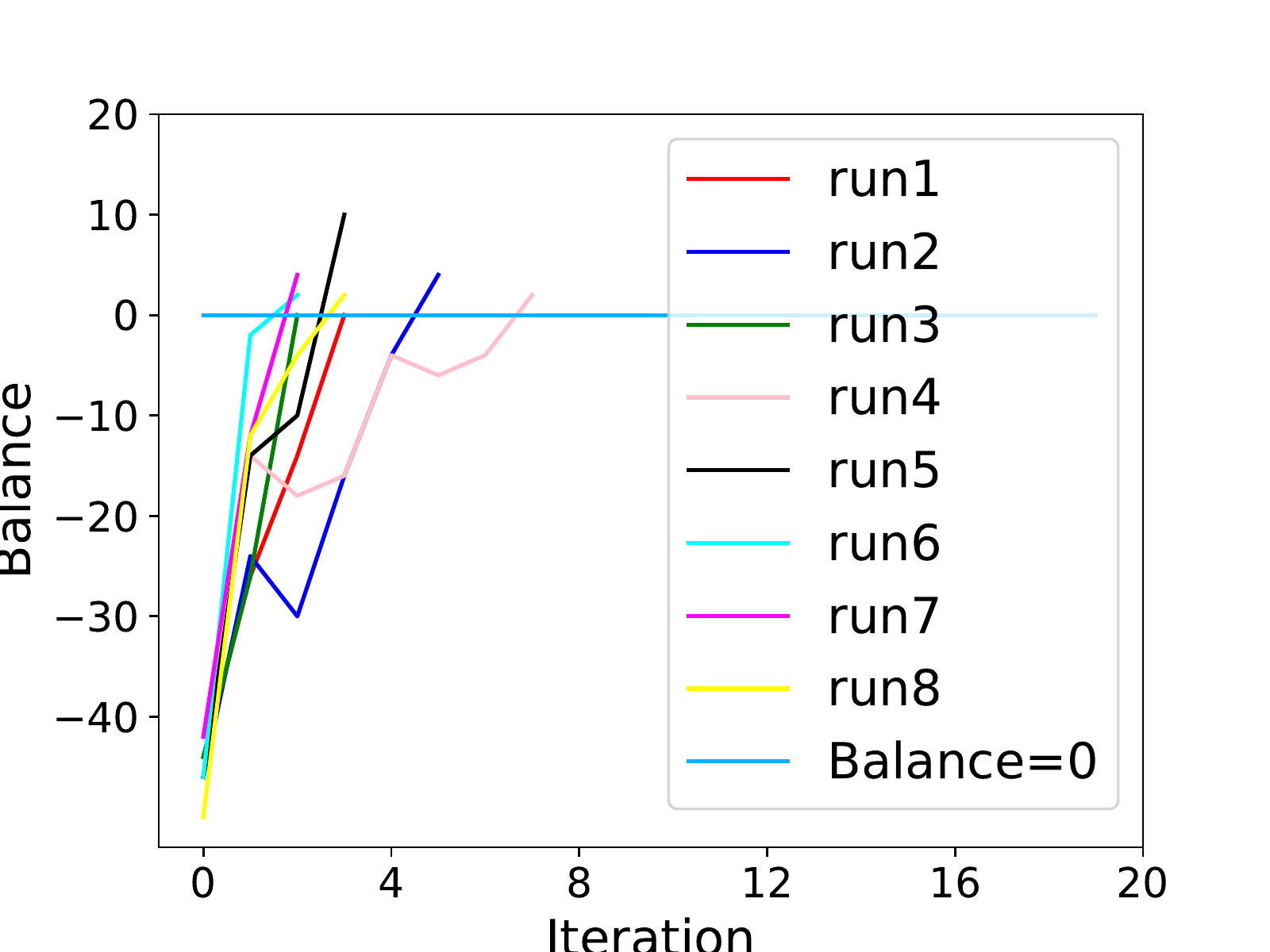}}
\hspace*{-1.5em}
\subfigure[WRoRa]{\label{fig:subfig:balancewroraothello}
\includegraphics[width=0.53\columnwidth]{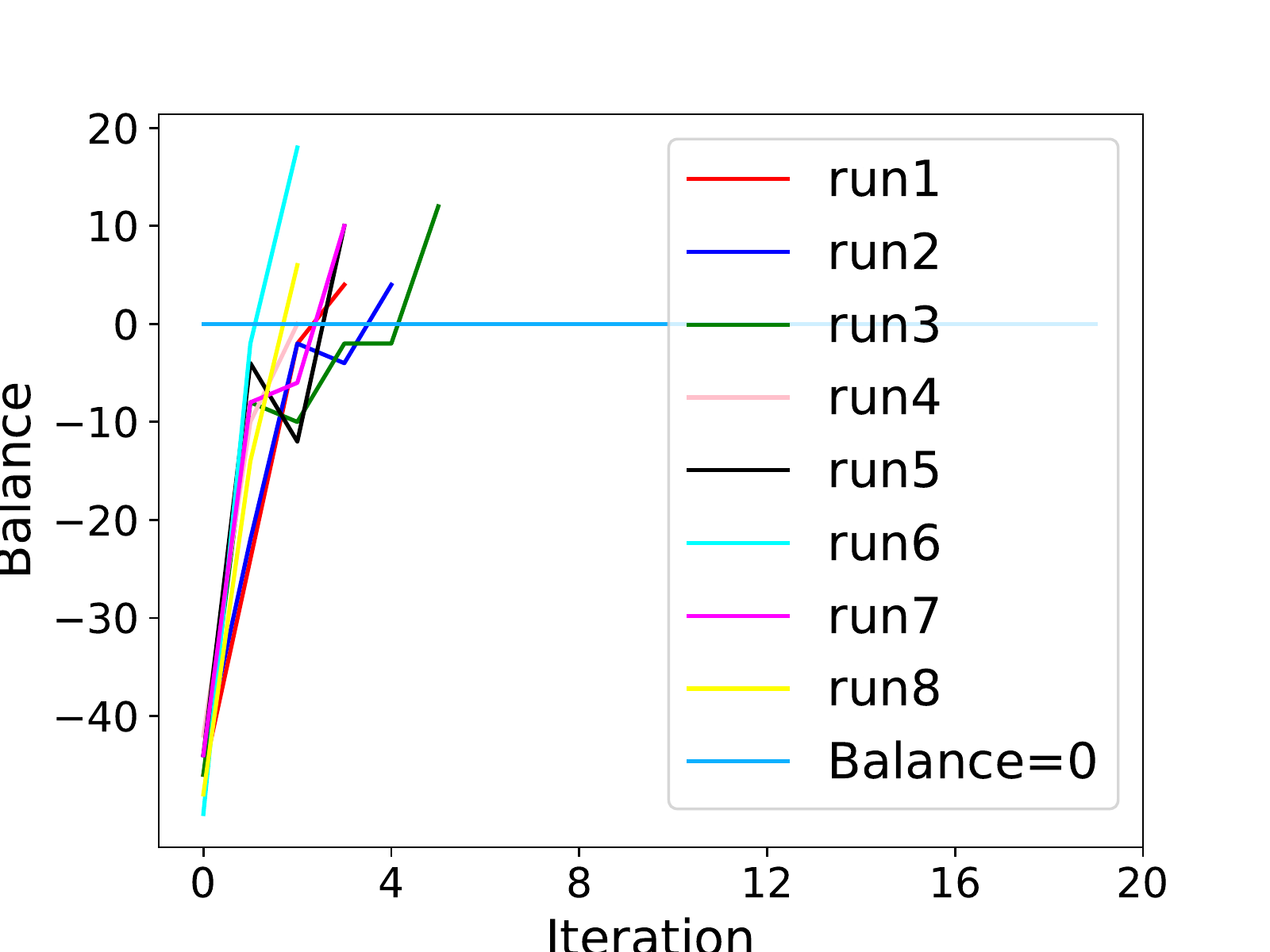}}

\caption{Reward balances of default MCTS while competing with different enhancements in self-play arena for 6$\times$6 Othello. Exceeding 0 means default MCTS defeats the enhancement, switch occurs.}
\label{fig:balanceofothello} 
\end{figure}

In addition, we show the arena results~(wins of default MCTS minus wins of enhancement) in each training iteration before switch happens in each run over 8 repetitions on Othello as an example in Fig~\ref{fig:balanceofothello}. In most curves, we can see improving reward balances achieved by default MCTS since it is getting stronger. 

\begin{figure*}[t!]
\centering
\subfigure[Connect Four]{\label{fig:adaptconnect4}
\includegraphics[width=0.8\columnwidth]{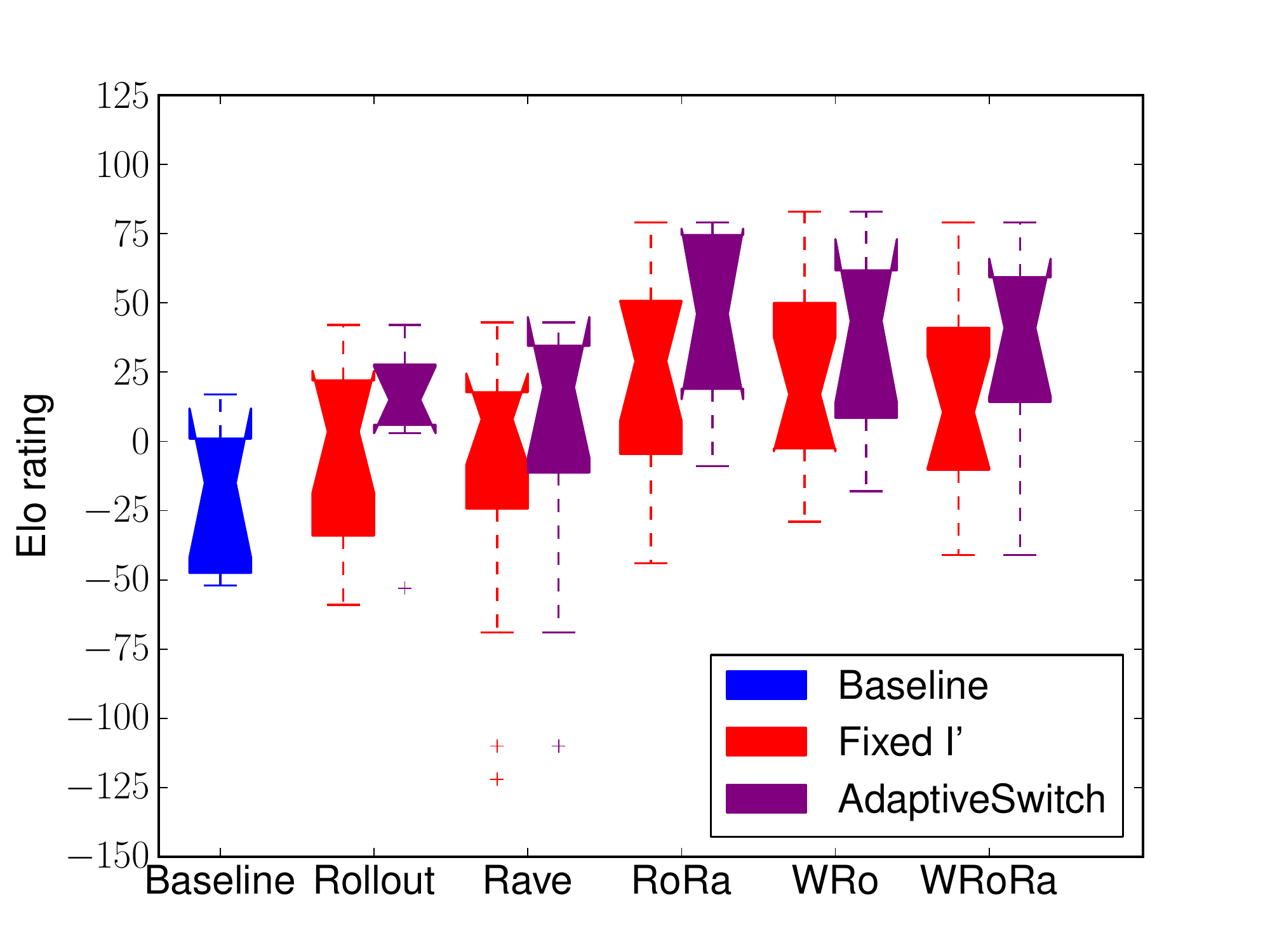}}
\subfigure[Othello]{\label{fig:adaptothello}
\includegraphics[width=0.8\columnwidth]{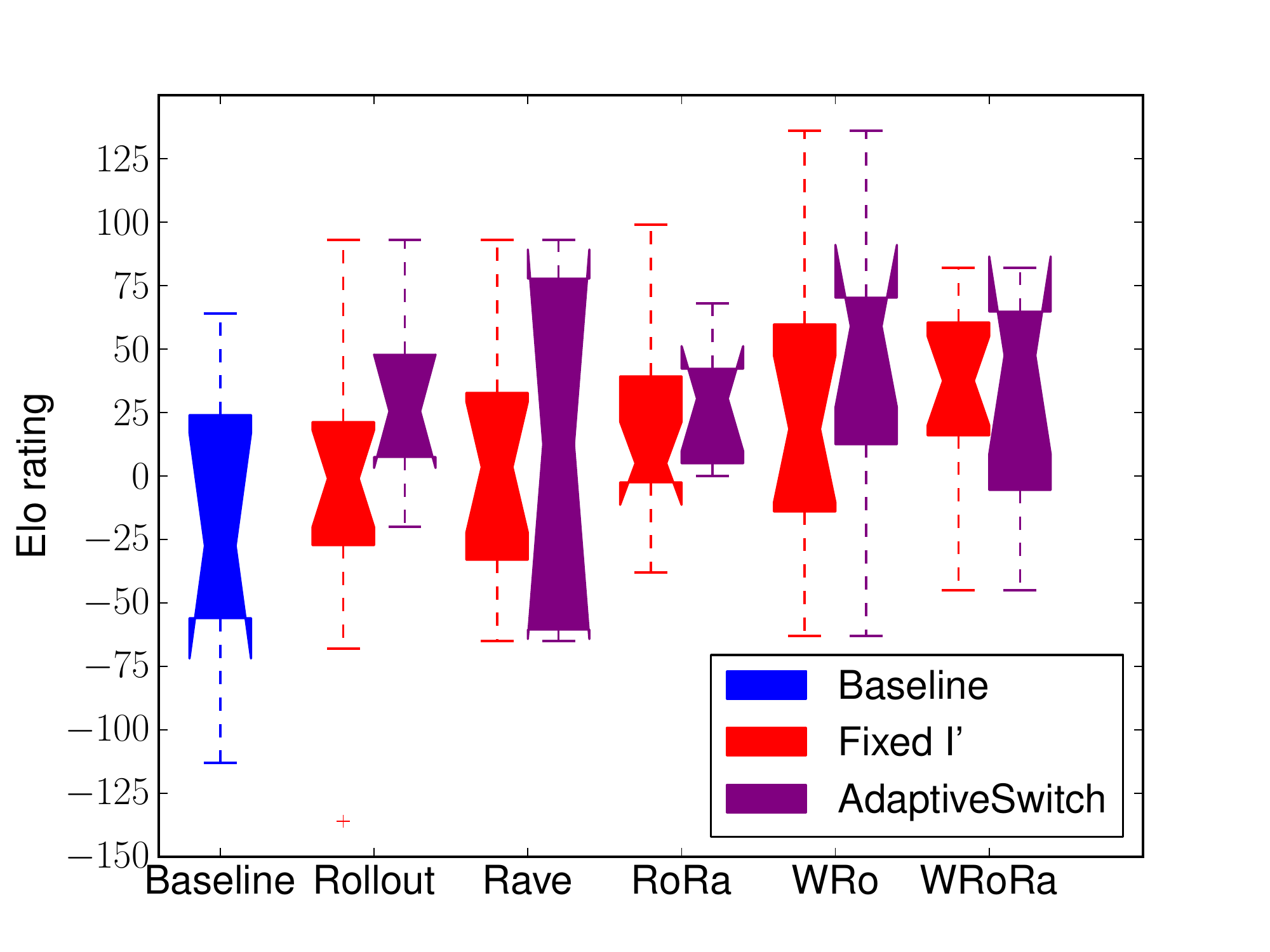}}
\caption{Comparison of adaptive switch method versus fixed $I^\prime$ based on a full tournament for 6$\times$6 Connect Four and Othello}
\label{fig:fixedvsadapt} 
\end{figure*}

More importantly, we collect all trained models based on our adaptive method, and let them compete with the models trained using fixed $I^\prime=5$ in a full round-robin tournament where each 2 players play 20 games. 

From Fig~\ref{fig:fixedvsadapt}, we see that, generally, on both Connect Four and Othello, all fixed $I^\prime$ achieve higher Elo ratings than the Baseline, which was also reported in~\cite{wang2020warm}). And all adaptive switch models also perform better than the Baseline. Besides, for each enhancement, it is important that the Elo ratings of the adaptive switch models are higher than for the fixed $I^\prime$ method, which suggests that our adaptive switch method leads to better performance than the fixed $I^\prime$ method when controlling the warm-start iteration length. Specifically, we find that for Connect Four, WRo and RoRa achieve the higher Elo Ratings~(see Fig~\ref{fig:adaptconnect4}) and  for Othello, WRoRa performs best~(see Fig~\ref{fig:adaptothello}), which reproduces the consistent conclusion~(at least one combination enhancement performs better in different games) as ~\cite{wang2020warm}).

In addition, for Connect Four, comparing the tuning results in Fig~\ref{fig:subfigdifferentheuristics} and the \emph{switch iterations} by our method in Table~\ref{tab:Ifordifferentgames}, we find that our method generally needs a shorter warm-start phase than employing a fixed $I^\prime$. The reason could be that in our method, there are always 2 different players playing the game, and they provide more diverse training data than a pure self-play player. In consequence, the  neural network also improves more quickly, which is highly desired.

Note that while we use the default parameter setting for training in the Gobang game, the \emph{switch} occurs at the first iteration. And even though we enlarge the simulation times for MCTS, only a few training runs shortly keep using the enhancements. We therefore presume that it is meaningless to further perform the tournament comparison for Gobang.

\section{Discussion and Conclusion}\label{sec:conclusion}

Since AlphaGo Zero' results, self-play has become a default approach for generating  training data tabula rasa, disregarding other information for training. However, if there is a way to obtain better training examples from the start, why not use them,  as has been done recently in StarCraft~(see DeepMind's AlphaStar~\cite{vinyals2019grandmaster}). In addition~\cite{wang2020warm} investigate the possibility of utilizing MCTS enhancements to improve AlphaZero-like self-play. They embed Rollout, RAVE and combinations as enhancements at the start period of iterative self-play training and tested this on small board games. Since the neural network and the MCTS statistics are initialized to random weights and zero, self-play suffers from a cold-start problem, and starting from scratch can lead to unstable learning at the start of the training. These problems can be cured by feeding human expert data or running MCTS enhancements or similar methods in order to generate expert data for training the neural network before switching to pure self-play.   (Not unlike RAVE warm-starts the winrate statistics of the original MCTS in 2007.)

Confirming~\cite{wang2020warm}, we find that finding an optimal value of fixed $I^\prime$ is difficult, therefore, we propose an adaptive method for deciding when to switch. We also use Rollout, RAVE, and combinations with network values to quickly improve MCTS tree statistics~(using RAVE) with meaningful information~(using Rollout) before we switch to Baseline-like self-play training. We employed the same games, namely the 6x6 versions of Gobang, Connect Four, and Othello.
In these experiments, we find that, for different games, and even different training runs for the same game, the new adaptive method generally switches at different iterations. This indicates the noise in the neural network training progress for different runs.
After 100 self-play iterations, we still see the effects of the warm-start enhancements as playing strength has improved in many cases, and for all enhancements, our method performs better than the method proposed in~\cite{wang2020warm} with $I^\prime$  set to 5. In addition, some conclusions are consistent to~\cite{wang2020warm}, for example, there is also at least one combination that performs better. 

The new adaptive method works especially well on Othello and Connect Four, "deep" games with a moderate branching factor, and less well on Gobang, which has a larger branching factor. In the  self-play arena, the default MCTS is already quite strong, and for games with a short and wide episode, the MCTS enhancements do not benefit much.  Short game lengths reach terminal states early, and  MCTS can use the true reward information more often, resulting in a higher chance of winning.
Since, Rollout still needs to simulate, with a limited simulation count it is likely to not choose a winning terminal state but a state that has the same average value as the terminal state. In this situation, in a short game episodes, MCTS works better than the enhancement with $T^\prime$=15. With ongoing training of the neural network, both players become stronger, and as the game length  becomes longer,  $I^\prime=5$ works better than the the Baseline. 

Our experiments are with small games. Adaptive warm-start works best in deeper games, suggesting a larger benefit  for bigger games with deeper lines. Future work includes larger games with deeper lines, and  
using   different but stronger enhancements to generate training examples.


\bibliographystyle{named}
\bibliography{ijcai21}

\end{document}